\documentclass[sigconf,nonacm]{acmart}
\usepackage{hyperref}       % hyperlinks
\usepackage{url}            % simple URL typesetting
\usepackage{booktabs}       % professional-quality tables
\usepackage{amsfonts}       % blackboard math symbols
\usepackage{nicefrac}       % compact symbols for 1/2, etc.
\usepackage{xcolor}         % colors
\usepackage{multirow}
\usepackage{listings}
\usepackage{pgfplots}
\usepackage{tcolorbox}
\usepackage{subcaption}
\usepackage{caption}
\usepackage{enumitem}
\usepackage{multirow}
\usepackage{parskip}
\usepackage{tikz-cd}
\usepackage{tikz}
\usepackage{csvsimple} % Required for importing CSV files
\usepackage{makecell}
\usepackage{parcolumns}
\usepackage{dirtytalk}

\setcopyright{none}
\usetikzlibrary{external}

\usepgfplotslibrary{groupplots}

\definecolor{codegreen}{rgb}{0,0.6,0}
\definecolor{codegray}{rgb}{0.5,0.5,0.5}
\definecolor{codepurple}{rgb}{0.58,0,0.82}
\definecolor{backcolour}{rgb}{0.95,0.95,0.92}
\lstdefinestyle{mystyle}{
    backgroundcolor=\color{white},   
    commentstyle=\color{codegreen},
    keywordstyle=\color{magenta},
    numberstyle=\tiny\color{codegray},
    stringstyle=\color{codepurple},
    basicstyle=\ttfamily\small,
    language=C,                                              
    numbers=none,                                      
    showspaces=false,                
    showstringspaces=false,
    showtabs=false,                  
    tabsize=2,
    breakatwhitespace=false,         
    breaklines=false,
    captionpos=b,
    frame=single,
    escapeinside={(*@}{@*)}
}
\pgfplotsset{compat=1.17}
 \usepackage{etoolbox}
\makeatletter
\patchcmd{\@verbatim}
  {\verbatim@font}
  {\verbatim@font\small}
  {}{}
\makeatother

\hypersetup{
   breaklinks=true,   % splits links across lines
   colorlinks=true,   % displays links as colored text instead of blocks
}

\definecolor{archtBlue}{RGB}{0, 114, 178}
\definecolor{archtRed}{RGB}{213, 94, 0}
\definecolor{archtGreen}{RGB}{0, 128, 0}
\definecolor{archtPurple}{RGB}{128, 0, 128}
\definecolor{archtOrange}{RGB}{230, 159, 0}
\definecolor{archtPink}{RGB}{204, 121, 167}

\newcommand{\hpcorpus}[0]{\textsc{HPCorpus}}

\newcommand{\comp}[0]{\textsc{MonoCoder}}

\newcommand{\hpcorpusomp}[0]{\emph{HPCorpus\textsubscript{OMP}}}

\newcommand{\ompify}[0]{\textsc{OMPify}}
\newcommand{\ompifygen}[0]{\textsc{OMPar}}
\newcommand{\mypara}[1]{\textit{#1}}

\makeatletter
\newcommand*{\rom}[1]{\expandafter\@slowromancap\romannumeral #1@}
\makeatother

\copyrightyear{2024}
\acmYear{2024}
\acmBooktitle{}
\acmDOI{}
\acmISBN{}

\author{Tal Kadosh}
\email{talkad@post.bgu.ac.il}
\affiliation{%
  \institution{Ben-Gurion University, IAEC}
  \country{Israel}
}

\author{Niranjan Hasabnis}
\email{niranjan@codemetal.ai}
\affiliation{%
  \institution{Code Metal}
  \country{United States}
}

\author{Prema Soundararajan}
\email{prema@uab.edu}
\affiliation{%
  \institution{University of Alabama at Birmingham}
  \country{United States}
}

\author{Vy A. Vo}
\email{vy.vo@intel.com}
\affiliation{%
  \institution{Intel Labs}
  \country{United States}
}

\author{Mihai Capotă}
\email{mihai.capota@intel.com}
\affiliation{%
  \institution{Intel Labs}
 \country{United States}
}

\author{Nesreen Ahmed}
\email{nesahmed@cisco.com}
\affiliation{%
  \institution{Cisco}
 \country{United States}
}

\author{Yuval Pinter}
\email{pintery@bgu.ac.il}
\affiliation{%
  \institution{Ben-Gurion University}
 \country{Israel}
}

\author{Gal Oren}
\email{galoren@stanford.edu}
\affiliation{%
  \institution{Technion, Stanford University}
 \country{United States}
}

\begin{document}

\title{\ompifygen{}: Automatic Parallelization with AI-Driven Source-to-Source Compilation}

\begin{abstract}
Manual parallelization of code remains a significant challenge due to the complexities of modern software systems and the widespread adoption of multi-core architectures. This paper introduces \ompifygen{}, an AI-driven tool designed to automate the parallelization of C/C++ code using OpenMP pragmas. \ompifygen{} integrates Large Language Models (LLMs) through two key components: \ompify{}, which assesses loop parallelization potential, and \comp{}-\textit{OMP}, a new fine-tuned model which generates precise OpenMP pragmas.

The evaluation of \ompifygen{} follows the same rigorous process applied to traditional tools like source-to-source AutoPar and ICPC compilers: (1) ensuring the generated code compiles and runs correctly in serial form, (2) assessing performance with the gradual addition of threads and corresponding physical cores, and (3) verifying and validating the correctness of the code’s output. Benchmarks from HeCBench and ParEval are used to evaluate accuracy and performance.

Experimental results demonstrate that \ompifygen{} significantly outperforms traditional methods, achieving higher accuracy in identifying parallelizable loops and generating efficient pragmas. Beyond accuracy, \ompifygen{} offers advantages such as the ability to work on partial or incomplete codebases and the capacity to continuously learn from new code patterns, enhancing its parallelization capabilities over time. These results underscore the potential of LLMs in revolutionizing automatic parallelization techniques, paving the way for more efficient and scalable parallel computing systems.

The sources of this work are available at our \textcolor{blue}{\href{https://github.com/Scientific-Computing-Lab/OMPar}{\ompifygen{}}} repository.
\end{abstract}

\maketitle

\section{Introduction}

In the last two decades, the computing industry has increasingly embraced multi-core and many-core shared memory architectures~\cite{intel2023hpccluster}. To meet the ever-growing demand for performance, computational paradigms have evolved to incorporate shared memory parallelism~\cite{kukanov2007foundations,buttlar1996pthreads,blumofe1995cilk}. OpenMP~\cite{mattson2019openmp} is a leading API that facilitates this model, which includes compiler directives, library routines, and environment variables. OpenMP enables parallel (multi-threaded) execution in a shared memory environment, harnessing the potential of modern multi-core and many-core systems. Its user-friendly and flexible interface has made it a popular choice for developers aiming to parallelize their applications.

Manual translation of serial code into parallel is a challenging and time-consuming task, often requiring in-depth knowledge of parallel programming APIs, paradigms, and an intricate understanding of program dependencies~\cite{marowka2010pitfalls}. Developers must identify parallelization opportunities, manage data dependencies, and ensure load balancing to maximize the performance of parallel code execution. Suffice to say, manual parallelization is prone to errors, and even minor oversights can lead to subtle bugs or performance degradation. Moreover, as software systems grow in complexity, manual parallelization becomes increasingly impractical, hindering the scalability and maintainability of parallel software.

% Automatic parallelization compilers~\cite{gnu,lattner2008llvm,icc} -- mostly using OpenMP -- offer a promising solution to alleviate the burden of manual parallelization by automatically identifying and exploiting parallelism in programs. These compilers analyze the program's structure and dependencies to identify loops and sections of code that can be executed concurrently. Automatic parallelization compilers aim to optimize program performance by distributing the workload across multiple processing units, thereby harnessing the computational power of parallel architectures.

Given the limitations of manual code parallelization, automatic parallelization was always an attractive idea. Automatic parallelization approaches can be broadly classified into two types: (1) formal approaches, and (2) AI-based approaches. Formal approaches for automatic parallelization use deterministic methods (such as syntax-driven translation rules etc). They can be further divided into source-to-source transformation~\cite{creusillet2009par4all,dave2009cetus,dever2015autopar} and parallelization during compilation~\cite{gnu,lattner2008llvm,icc}. Source-to-source transformation tools translate the program's source code into a parallelized version, typically by inserting parallelization directives or annotations (e.g., OpenMP pragmas) into the code. These directives guide a standard compiler (such as GCC or LLVM) in generating parallel execution plans. The effectiveness of source-to-source parallelization heavily relies on the accuracy of parallelization annotations and the compiler's ability to interpret them~\cite{prema2017identifying,harel2020source}. In contrast, auto-parallelizing compilers integrate parallelization techniques directly into the compiler's optimization pipeline, enabling the compiler to analyze and parallelize code at a lower level of abstraction~\cite{doerfert2018compiler}. While auto-parallelization compilers offer the advantage of seamless integration with existing compiler infrastructure, they often struggle with complex program structures and dynamic execution patterns, limiting their effectiveness in achieving optimal parallel performance~\cite{mansky2014specifying,olabi2022compiler}.

Recently breakthroughs in AI, such as large language models for natural language processing (NLP) and code generation have led to considerable interest in developing specific AI models for automatic code parallelization~\cite{chen2023lm4hpc, kadosh2023pragformer, mahmud2023autoparllm}. LLMs such as GPT-4 and LLaMA-3.1, on the other hand, are trained on vast amounts of open-source code and seem capable of parallelizing serial code (using variety of parallelization APIs such as CUDA, OpenMP, etc.) Yet when Nichols et al. recently evaluated popular LLMs with their ParEval benchmark~\cite{nichols2024can}, they found that LLMs struggle with code parallelization.
%Recent work by Nichols et al.~\cite{nichols2024can} that evaluated parallel programming capabilities of popular LLMs found that, in fact, LLMs seem to struggle with the code parallelization problem. 
Specifically, they conclude their findings by saying that \say{\textit{LLMs are significantly worse at generating parallel code than they are at generating serial code}}. They further comment that \say{\textit{the poor performance of LLMs on ParEval indicates that further efforts are necessary to improve the ability of LLMs to model parallel code and/or create new LLMs that are specialized for parallel code generation.}}

% Despite the advancements in automatic parallelization technology, existing compilers still face several challenges. First, the effectiveness of automatic parallelization heavily depends on the program's structure and the availability of parallelism at the loop or task level. Programs with irregular loop structures or complex control flow may not benefit significantly from automatic parallelization, leading to suboptimal performance. Second, automatic parallelization compilers often struggle with data dependencies and synchronization overhead, especially in shared-memory parallelization models. Ensuring data consistency and avoiding race conditions in parallelized code requires sophisticated analysis and runtime support, adding complexity and overhead to the parallelization process. Third, the performance of automatic parallelization compilers may be limited by hardware constraints and architectural bottlenecks. Inefficient memory access patterns, cache coherence overhead, and communication latencies can impact the scalability and efficiency of parallel execution.

Our paper proposes a novel automatic parallelization approach that addresses the limitations of existing auto-parallelization approaches (discussed in detail in \autoref{background}). Our approach, named \ompifygen{}, employs a modular approach by integrating multiple state-of-the-art AI models in the parallelization pipeline, each designed for a specific task -- \ompify{} for assessing the parallelization potential of loops and \comp{}-\textit{OMP} for generating precise OpenMP pragmas. This modular approach allows for greater flexibility and accuracy compared to monolithic LLMs that attempt to handle all aspects of code generation within a single model and can output a wide range of responses. Specifically, this approach allows us to evaluate every model individually with respect to its task (as it is typically done in an ablation study), enabling the possibility of fine-tuning them with specific datasets. In fact, \comp{}-\textit{OMP} is a domain-specific (HPC-specific) model that precisely exploits this capability by leveraging C/C++ serial and parallel programs for fine-tuning, thereby improving its performance on the task of parallelization and pragma generation.

%The precise fine-tuning of these models ensures that we can rely more confidently on their responses. By harnessing the advanced capabilities of domain-specific language models, \ompifygen{} aims to overcome the limitations of existing automatic parallelization compilers and deliver superior performance across a wide range of applications. While existing LLM-based approaches have shown promise, they often fall short in several critical areas. First, many LLMs, although trained on vast datasets that include some HPC code, are not specifically tailored to the intricacies of parallel programming paradigms such as OpenMP. In contrast, domain-specific models like those employed in \ompifygen{} are trained exclusively on C/C++ and high-performance code, which significantly elevates the predictability and reliability of the AI model for our specific needs. 

%\niranjan{I will add a line about why we have two models and ablation study}
%\niranjan{This paragraph looks a bit weak to me in the context of prior paragraphs. Comparison with formal tools is fine - we are covering that in the background; for LLMs though, we are not saying why our approach is better than existing LLMs - is there anything fundamental?}

\mypara{\textbf{Contributions.}} This paper makes following contributions:

\begin{itemize}[left=0pt]
    \item We propose \ompifygen{}, a novel approach to automatic parallelization that leverages smaller but efficient and more accurate domain-specific models for loop parallelization (\ompify{}~\cite{kadosh2023advising}) and code generation (\comp{}~\cite{kadosh2023domain} fine-tuned on OpenMP codes, \\ \comp{}-\textit{OMP}).
    \item We evaluate the accuracy of \ompifygen{} in generating accurate OpenMP pragmas. We also measure the performance of the generated pragmas using ground-truth labels and compile-and-run checks against the HeCBench and ParEval benchmarks. These are compared with automatic parallelization compilers (ICPC and AutoPar).
    \item Our results show that \ompifygen{} outperforms both ICPC and AutoPar in suggesting accurate parallelization pragmas. Moreover, our results demonstrate that more than 90\% of \ompifygen{} suggested loops pass both compilation and functional tests, suggesting that \ompifygen{} understands the syntax and semantics of the parallelization pragmas. Overall, results demonstrate that \ompifygen{} achieves superior performance compared to existing auto-parallelization compilers, even when provided with partial codes, highlighting the robustness and scalability of our approach.
\end{itemize}

By addressing the limitations of current automatic parallelization techniques and leveraging the power of AI-driven parallelization, \ompifygen{} represents a significant advancement in the field of automatic parallelization, offering a promising solution for efficiently harnessing parallelism in modern software systems.

\section{Background and Motivation}
\label{background}
In this section, we discuss both formal as well as AI-based approaches for automatic code parallelization. Moreover, we present concrete examples of the limitations of these approaches as a motivation for our approach.

\begin{figure*}[ht!]
\begin{minipage}[t]{.33\textwidth}
\begin{lstlisting}[caption={Example 1}]
size_t findLastShortBook(
  const class std::vector<Book>,
  std::allocator<Book>> &books) {
(*@ \textcolor{orange}{\#pragma omp parallel for} @*)
 for (size_t i = books.size();
      i >= ((unsigned long )0) + 1;
      i += -1) {
   if (books[(i-1)].pages < 100) {
     return i - 1;}}
 // If no book with pages < 100 is
 // found, return an appropriate
 // value (e.g., books.size()).
 return books.size();
 }
\end{lstlisting}
\end{minipage}
\hfill
\begin{minipage}[t]{.32\textwidth}
\begin{lstlisting}[caption={Example 2}]
size_t findFirstEven(
  const class std::vector<int,
  std::allocator<int>> &x) {
(*@ \textcolor{orange}{\#pragma omp parallel for} @*)
 for (size_t i = 0;
      i <= x.size() - 1;
      i += 1) {
   if (x[i] % 2 == 0) {
    // Check if current element
    // is even. Return the index
    // if found
    return i;}}
 // Return -1 otherwise.
 return (-1);}


\end{lstlisting}
\end{minipage}
\hfill
\begin{minipage}[t]{.3\textwidth}
\begin{lstlisting}[caption={Example 3}]
int edgeCount(
  std::vector<int> const& A,
  size_t N) {
 int count = 0;
 size_t i = 0, j = 0;
(*@ \textcolor{orange}{\#pragma omp parallel for \\} @*)
  (*@ \textcolor{orange}{reduction(+:count)} @*)
 for (i = 0; i < N; ++i) {
   for (j = 0; j < N; ++j) {
     if (A[i * N + j] == 1) {
       #pragma omp critical
       count++;}}}
 return count;
 }
\end{lstlisting}
\end{minipage}
\vspace{-0.1in}
\caption{Examples of incorrect parallelization by AutoPar: The first two examples show \texttt{for} loops that were parallelized by AutoPar but the parallelization changed the loop semantics. The last example shows a loop that can be parallelized but AutoPar missed the parallelization opportunity.}
\label{fig:autopar_incorrect_parallelization}
\end{figure*}

\subsection{Formal Approaches to Automatic Parallelization}

To mitigate the difficulties of manual parallelization, various automatic parallelization tools have been developed to assist programmers in automatically converting sequential programs into parallel ones. These tools free programmers from the need to manually insert parallelization directives, thus simplifying the software development process. These tools typically focus on loops, as loops are where programs spend the majority of their execution time. 

Broadly, auto-parallelization tools can be classified into two types: compilers and source-to-source transformation tools.

Given that automatic parallelization is a form of program transformation, popular compilers like GCC~\cite{gcc_openmp_support}, LLVM~\cite{llvm_autoparallel}, and ICC~\cite{icc_autoparallel} also perform automatic parallelization. For instance, GCC has supported automatic parallelization since version 4.3, released in 2012, and LLVM's Polly~\cite{grosser2012polly,ouellet2016parallelisation} provides loop optimization and parallelization accessible through Clang~\cite{lattner2008llvm}. These compilers leverage existing program analysis infrastructure, such as data-flow analysis, to ensure the correctness of the parallelization. They also use heuristics to prevent performance degradation by ensuring that parallelized loops have sufficient iterations.

An alternative approach to compiler-based parallelization is the source-to-source (S2S) transformation. S2S tools convert sequential code into parallel code while maintaining the original source code, allowing subsequent compilation with any standard compiler. Examples of S2S tools include AutoPar~\cite{dever2015autopar}, Par4all~\cite{amini2012par4all}, and Cetus~\cite{dave2009cetus}, or a combination of those with ComPar~\cite{mosseri2020compar}. S2S tools have the advantage of keeping the source code clean and enabling the use of different compilers for further optimization and deployment.

Despite their advantages, automatic parallelization tools face significant challenges~\cite{prema2017identifying,milewicz2021negative, prema2019study}. One major issue is the substantial manual effort required to develop and maintain these tools. They are typically rule-based and rely on pattern matching, which necessitates continuous updates to keep up with revisions to the OpenMP specification~\cite{ompcompilers}. For instance, the latest version of GCC does not fully support all features of OpenMP v5.0, and many S2S tools fail to incorporate newer OpenMP capabilities, such as task-based parallelism and offloading kernels to devices. Moreover, automatic parallelization tools can be overly strict, often failing to recognize opportunities for parallelization~\cite{harel2020source}. This conservativeness, while ensuring correctness, can lead to missed opportunities where parallelization could have been safely and beneficially applied. Additionally, when these tools do perform parallelization, they can sometimes result in suboptimal performance scalability~\cite{harel2020source}. The resulting parallelized code might not efficiently utilize the available computational resources, leading to limited performance gains or even performance degradation in some cases.

\subsubsection{\textbf{A motivating case study of the limitations of formal auto-parallelizing tools.}}
\label{section:motivation}

To concretely assess the limitations of these formal tools, we decided to conduct a case study of two state-of-the-art auto-parallelization tools in their ability to parallelize serial programs. Specifically, we chose AutoPar and ICPC. AutoPar~\cite{liao2010semantic} is a state-of-the-art static source-to-source-based automatic parallelization tool. ICPC is the Intel C++ compiler, which provides advanced auto-parallelization capabilities as part of high-performance computing tools. ICPC~\cite{icpc} performs automatic parallelization at compile-time, leveraging both static analysis and dynamic profiling data to optimize code for modern multi-core architectures. %Unlike AutoPar, which focuses on OpenMP directive insertion, ICPC integrates more advanced optimizations such as vectorization and memory access improvements, targeting low-level hardware features.

\mypara{Test dataset.} As these tools would have been evaluated on existing serial-code benchmarks, we decided to generate a synthetic benchmark of serial code. Specifically, we decided to leverage ParEval~\cite{nicholsCanLargeLanguage2024}, the latest work that has developed a benchmark to evaluate the ability of LLMs to generate parallel programs. Specifically, ParEval contains LLM prompts that represent 60 different coding problems related to scientific and parallel computing, with each problem, solved using programs written in 7 different parallel programming languages (such as CUDA, OpenMP, etc.) Moreover, every program has pre-defined test inputs as well as expected outputs. We collected 60 OpenMP-based programs generated by GPT-3.5-turbo and converted these parallel programs into serial ones by removing all OpenMP pragmas\footnote{For the pragma erasing, we used AutoParBench's PragmaRemover tool: \url{https://github.com/LLNL/AutoParBench/tree/master/tools/PragmaRemover}}. We found that these programs were using the C++17 standard.

\mypara{Evaluation methodology.} Both AutoPar and ICPC operate on complete programs (as opposed to a single \texttt{for} loop) for automatic parallelization. Consequently, we applied both tools to parallelize each and every serial program from the test set, attempted to compile generated OpenMP-parallelized programs, run with the test inputs (if the compilation was successful), and compared their output with the expected output. 

\mypara{Findings.} We found that ICPC could parallelize 2 of these 60 programs and 5 \texttt{for} loops in total. Similar to ICPC, AutoPar was able to parallelize 2 programs and 2 loops in total. When subjected to compile and run tests, ICPC-parallelized programs timed out, while AutoPar-parallelized programs encountered errors. We found that one of the reasons for the inability of these tools to parallelize input programs was possibly their incomplete support for the C++17 standard that is used by the input programs. We believe this limitation points to the need for manual efforts to support these tools.

We present three concrete examples of Autopar's incorrect parallelization in \autoref{fig:autopar_incorrect_parallelization}. Specifically, the first two examples are of incorrect parallelization (false positive) --- \texttt{for} loops that cannot be parallelized as per OpenMP specification --- while the third example is a missed parallelization opportunity (false negative). In particular, the first two examples contain \texttt{for} loops that have a return statement and the behavior of the loops is strictly based on the order of loop iterations. Parallelizing such a loop provably changes the semantics of the loop from its serial version. The third example contains a \texttt{for} loop that can be parallelized with OpenMP's \texttt{reduction(+:count)} clause, but AutoPar missed the parallelization opportunity. The missed opportunity could be because of non-affine constructs (i.e., \texttt{A [ i * N + j]}) in the loop body that AutoPar could not analyze statically. To summarize, we found that existing formal tools for auto-parallelization suffer from several limitations. Later in the evaluation, we present several more examples of incorrect parallelizations by AutoPar and ICPC.

%Another fundamental limitation is these tools' inability to handle incomplete or in-development code. They are designed to work on complete, compilable programs and depend on standard compiler passes like parsing and semantic analysis to evaluate the potential for parallelization. This limitation makes them unsuitable for use during the early stages of software development, where the code may not yet be fully developed or compilable.

\begin{table*}
    \centering
\begin{tabular}{|p{1.6cm}|p{1.2cm}|p{0.6cm}|p{3.2cm}|p{1.8cm}|p{1.7cm}|p{2.5cm}|p{1cm}|}
\hline
\textbf{Authors} & \textbf{Month} & \textbf{Year} & \textbf{Model} & \textbf{Task} & \textbf{Usage} & \textbf{Dataset} & \textbf{Ref} \\ \hline
Harel & April & 2022 & PragFormer & Classification & Fine-tuning & OpenOMP dataset & \cite{harel2023learning, kadosh2023pragformer} \\ \hline
Chen & May & 2023 & Graph2Par & Classification & Fine-tuning & OMPSerial dataset & \cite{chen2023learning} \\ \hline
Kadosh & May & 2023 & \ompify{} & Classification & Fine-tuning & HPCorpus dataset & \cite{kadosh2023advising} \\ \hline
Chen & June & 2023 & LM4HPC & Classification & Inference & OMP4Par dataset & \cite{chen2023lm4hpc} \\ \hline
Godoy & June & 2023 & Codex & Generation & Inference & Numerical Kernels & \cite{godoy2023evaluation} \\ \hline
Nichols & June & 2023 & HPC-Coder & Generation & Fine-tuning & HPC-Coder dataset & \cite{nichols2023modeling} \\ \hline
Valero-Lara & September & 2023 & Llama-2, GPT-3 & Generation & Inference & Numerical Kernels & \cite{valero2023comparing} \\ \hline
Mahmud & October & 2023 & AutoParLLM & Classification & Fine-tuning, Inference & OMPSerial dataset & \cite{mahmud2023autoparllm} \\ \hline
Pornman' & November & 2023 & CodeT5-FT & Classification & Fine-tuning & BigQuery public dataset & \cite{pornmaneerattanatri2023parallelizable} \\ \hline
Kadosh & December & 2023 & \comp{} & Generation & Pre-training & HPCorpus dataset & \cite{kadosh2023domain} \\ \hline
Nichols & January & 2024 & CodeLlama, StarCoder, GPT-3.5, GPT-4 & Generation & Inference & ParEval & \cite{nichols2024can} \\ \hline
Chen & January & 2024 & OMPGPT & Generation & Pre-training, Fine-tuning & HPCorpus dataset & \cite{chen2024ompgpt} \\ \hline
Rosas & June & 2024 & GPT-4, CodeLlama-70B & Generation & Inference & NAS, PolyBench & \cite{rosas2024should} \\ \hline
\end{tabular}
    %\csvautotabular{parallelmodels.csv}
    \caption{Comparison of LLM-based models for OpenMP code, showing their usage (Inference, Fine-tuning, Pre-training + Fine-tuning) and corresponding training data. Almost all of the models use common ML metrics for success, especially OpenMP pragma string-based comparison.
    \vspace{-0.7cm}}
    \label{table:csv_table}
\end{table*}

\subsection{AI and Large Language Models for Parallel Program Generation}

The emergence of large language models (LLMs), particularly those built on transformer architectures like the GPT (Generative Pre-trained Transformer) series~\cite{gpt}, has sparked a revolution in natural language processing (NLP). These models have exhibited exceptional prowess in comprehending and producing human language~\cite{devlin2018bert, liu2019roberta, raffel2020exploring}. Leveraging the inherent similarities between code and natural language, researchers have extended these capabilities to the realm of programming, paving the way for automating various software development tasks, including parallelization.

AI-based tools for automatic parallelization manifest in diverse forms, each offering a distinct approach and emphasis. OpenMP-specific tools address the challenge of OpenMP parallelization, scrutinizing serial code and proposing suitable OpenMP pragmas. Examples encompass PragFormer~\cite{harel2022learning}, \ompify{}~\cite{kadosh2023advising}, Graph2Par~\cite{chen2023learning}, HPCoder~\cite{kadosh2023domain}, and AutoParLLM~\cite{mahmud2023autoparllm}. On the other hand, pre-trained HPC-oriented models such as \comp{}~\cite{kadosh2023domain,kadosh2023scope} and OMP-GPT~\cite{chen2024position} are initially trained on expansive datasets before being fine-tuned for OpenMP-related tasks, capitalizing on their expansive comprehension of code structures and parallelization principles. We capture comparative details of these AI-based automatic parallelization approaches in \autoref{table:csv_table}, which highlights the diverse methodologies used in the field. These models differ not only in their underlying architectures but also in the nature of the tasks they are designed for and the datasets they leverage.

Meanwhile, general-purpose LLMs such as GPT-4~\cite{openaichatgpt} and CodeLLaMa~\cite{roziere2023code} are versatile and capable of addressing a range of programming tasks, including OpenMP parallelization. However, they lack specialized training in parallel programming paradigms. Due to the limited exposure to these paradigms in their training data, these models often struggle with key parallel programming challenges, such as reasoning about data distribution, managing race conditions, and implementing complex parallel algorithms~\cite{nichols2024can}.

%Critical design choices impact the efficacy of AI-based parallelization tools. The formulation of the problem often entails decomposing the task into subproblems, such as assessing parallelization potential and recommending suitable pragmas. The representation of source code is paramount for precise analysis, with methodologies like abstract syntax trees (ASTs) and data-flow graphs (DFGs) facilitating the capture of code structures and dependencies. Model architecture (e.g., transformer, graph neural networks) is another crucial decision for AI researchers. Finally, training the models benefits from synthesizing datasets from diverse sources. 

Performance evaluations have previously demonstrated the superiority of AI-based approaches over traditional methods. Notably, PragFormer has surpassed the source-to-source tool ComPar~\cite{mosseri2020compar} in discerning parallelization potential. Similarly, AI-based tools have been advancing rapidly. Graph2Par has exhibited greater precision in predicting applicable OpenMP clauses compared to PragFormer. Moreover, both \ompify{} and PragFormer have outperformed ChatGPT in discerning the parallelization potential of loops.

However, despite these advancements, a significant gap remains in matching the capabilities of automatic parallelization compilers. Unlike previous AI-specific metrics that primarily focused on next-token prediction accuracy and comparison to ground truth, automatic parallelization tools are traditionally evaluated based on machine performance metrics such as runtime, scalability, and accuracy of computation results. \ompifygen{} addresses this challenge by validating performance using HPC-centric methodologies as recommended by ParEval~\cite{nichols2024can}.

\begin{figure*}[!ht]
\centering
\includegraphics[width=\textwidth]{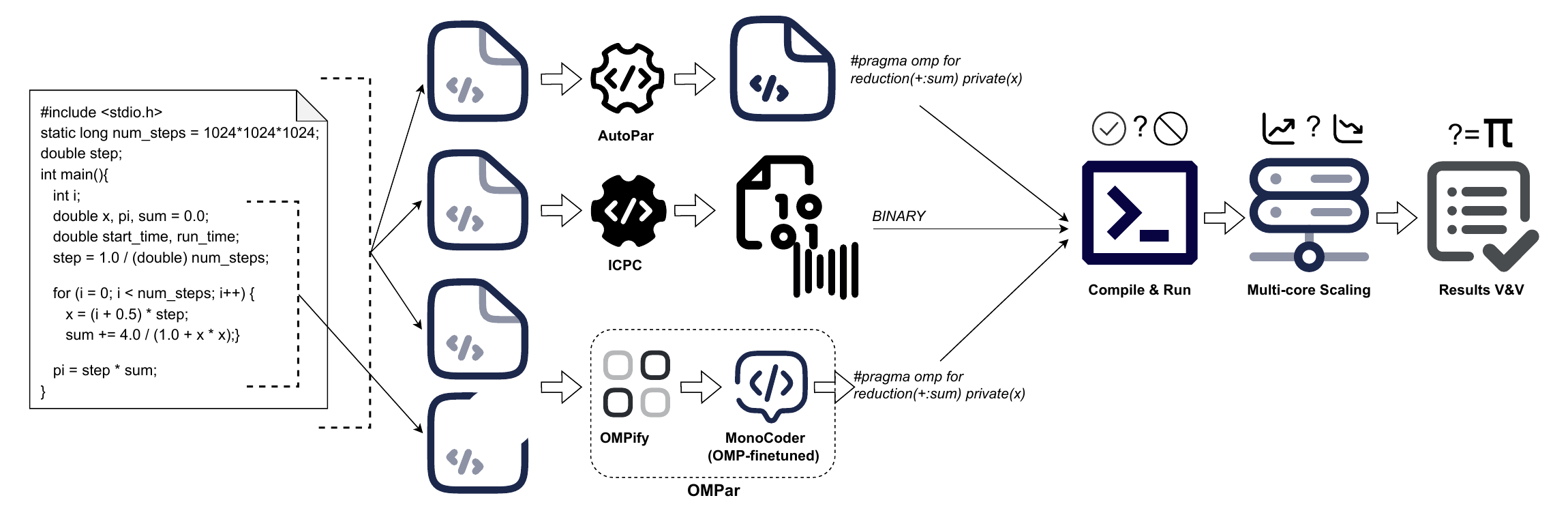} 
\caption{\ompifygen{} workflow, given a simple pi code example, and comparison with other compilers: While source-to-source automatic compilers (such as AutoPar) determine the need and generate the needed pragma, other HPC compilers (such as ICPC) can generate a binary that is already parallel. Nevertheless, both are rule-based and need to compile the code. In sharp contrast, \ompifygen{} is based on two dedicated LLMs, one trained for classifying the need for parallelization (\ompify{}) and one for generating the full pragma (\comp{}-\textit{OMP}). Both models were trained on \hpcorpus{} and \hpcorpusomp{}, a large-scale collection of codes from GitHub. The evaluation of AutoPar, ICPC, and \ompifygen{} is the same: (1) check whether the generated code compiles and runs serially, (2) evaluate their performances with the gradual addition of threads and corresponding physical cores, and (3) finally verify and validate the code's output.}
\label{fig:teaser}
\end{figure*}

\subsubsection{\textbf{Limitations of LLMs in auto-parallelization.}}

%We have covered comparative analysis of existing parallelization-specific AI models in \autoref{table:csv_table}. We now talk briefly about the limitations of LLMs in automatic parallelization. 
Several recent papers~\cite{nichols2024can} have systematically evaluated the ability of LLMs to generate parallel code. In particular, authors of ParEval~\cite{nichols2024can} have presented a systematic analysis of the ability of various popular LLMs, including GPT-4, in auto-parallelization. As such, we quote the findings from these studies to summarize: \say{\textit{LLMs are significantly worse at generating parallel code than they are at generating serial code}}. Moreover, they further comment that \say{\textit{the poor performance of LLMs on the ParEval benchmark indicates that further efforts are necessary to improve the ability of LLMs to model parallel code and/or create new LLMs that are specialized for parallel code generation.}}

One major limitation contributing to the poor performance of current LLMs in parallelization tasks is that these models are primarily trained on raw source code, without leveraging the deeper code representations used by compilers, such as Abstract Syntax Trees (AST), Control Flow Graphs (CFG), and Data Flow Graphs (DFG). While source code provides surface-level syntax and semantics, it lacks the structural insights necessary for complex tasks like parallelization, which require understanding dependencies, data flow, and control flow. Compiler-level representations like AST capture the hierarchical structure of code, while DFGs model how data moves through variables and computations, both of which are crucial for identifying independent tasks that can be parallelized. The absence of these representations limits the ability of LLMs to detect parallelization opportunities, manage data dependencies, and ensure race condition-free execution, making them less effective in generating optimized parallel code.

\section{\ompifygen{}: LLM-based Source-to-Source Parallelization Compiler}
\label{openm}

This section presents the full pipeline for source-to-source automatic parallelization using our AI-driven tool \ompifygen{}. \autoref{fig:teaser} shows the high-level overview of the pipeline.

\mypara{Task objective.} While a lot of progress has been made in integrating ML into the identification of parallelism opportunities in C/C++ programs~\cite{kadosh2023advising, kadosh2023pragformer, shen2021towards}, there has been a notable gap in \textit{connecting parallelism detection with code generation}. To bridge this gap, our work leverages \ompify{} for parallelism detection and \comp{} for parallel code generation, refining its capabilities through fine-tuning the OpenMP-specific subcorpus of \hpcorpus{}. Strongly influenced by the traditional pipeline compilation of source-to-source compilers, we created \ompifygen{} by pipe-lining \ompify{} with the fine-tuned \comp{} LM (henceforth \comp{}-\textit{OMP}) to generate complete OpenMP pragmas when presented with a \texttt{for} loop written in C/C++. OpenMP pragma annotations on \texttt{for} loops allow such loops to improve performance by using multi-threading to process computations inside the loop concurrently. 

\mypara{Task design.}  Instead of using static code analysis and applying rule-based methods and heuristics, \ompifygen{} first uses \ompify{} to determine whether the loop should have a pragma. If the pragma is not applicable, \ompifygen{} will return this result. If the pragma is deemed necessary, \comp{} is then used to generate the complete pragma for the given \texttt{for} loop. 

\mypara{Components of \ompifygen{}.} The two major components of \ompifygen{} are \ompify{} and the fine-tuned version of \comp{} on \hpcorpusomp{}, named \comp{}-\textit{OMP}. We chose those two out of \autoref{table:csv_table} as they are (1) SOTA models with excellent performance for their tasks with C and C++ languages, (2) trained on the same large-scale parallel-computing-oriented dataset, and (3) designed as targeted small domain-specific language models, which can be deployed even on average computer systems, similarly to the installation of AutoPar and ICPC.

\subsection{\ompify{}}

The first major component of \ompifygen{} is \ompify{}. \ompify{} is an encoder-only transformer model trained on \texttt{for} loops in C and C++ to classify whether OpenMP is needed or applicable. Specifically, it was trained on approximately 54K \texttt{for} loops, with about half of them containing OpenMP and the rest not. The \ompify{} pipeline operates as follows: First, it generates the given code's data flow graph (DFG). Then, the model is fed with multiple code modalities, including the DFG and the plain code. The model returns a binary classification for the detection task. Results demonstrate that \ompify{} outperforms existing approaches, such as the general-purpose and popular ChatGPT (GPT-3.5) and the targeted PragFormer models, in terms of F1 score and accuracy. Specifically, \ompify{} achieves up to 90\% accuracy on commonly used OpenMP benchmark tests such as NAS, SPEC, and PolyBench.

\subsection{\comp{}-\textit{OMP}}

The second major component of \ompifygen{} is the fine-tuned version of \comp{}, \comp{}-\textit{OMP}. \comp{} is a decoder-only model initially trained on corpora of HPC-programming languages. We fine-tuned it on approximately 25K \texttt{for} loops so that for a given \texttt{for} loop, \comp{} generates the corresponding OpenMP pragma, particularly the \textit{"\#pragma omp for"} along with private and reduction clauses. For evaluation, \comp{} was compared to the general-purpose ChatGPT (GPT-3.5) on the task of OpenMP pragma generation. \comp{} significantly outperformed ChatGPT in both predicting the correct clauses and using the relevant variables inside the clauses.

\mypara{Dataset.}
To create a dataset for the automatic OpenMP parallelization generation task, we curated a subcorpus of \hpcorpus{} named \hpcorpusomp{}. This subcorpus specifically includes all the \texttt{for} loops that have either \texttt{\#omp parallel for} pragma that distributes the workload across multiple CPU cores or offloads tasks to a team of threads for GPU execution using \texttt{\#omp target teams distribute}. The dataset also includes OpenMP pragmas containing \texttt{private}\footnote{To avoid the loss of crucial details, we designed \hpcorpusomp{} such that the \texttt{private} clause encompasses instances of both \texttt{firstprivate} and \texttt{lastprivate}, as both of these clauses serve the purpose of creating a private instance of variables for each thread.} and \texttt{reduction} clauses. It is important to note that we excluded the context of the loops to maintain focus on the loop-specific information. The details of the dataset in terms of these clauses are presented in \autoref{tab:omp_corpus}.

\begin{table}[!htbp]
\centering
\begin{tabular}{c||r|r|r|r}
\hline
\centering
                 & \texttt{private} & \texttt{reduction} & \texttt{target} & \texttt{parallel for}\\ \hline
C       & 3,526        & 713             & 145           & 8,764           \\ \hline
C++     & 2,122       & 1,424             & 345          & 18,151          \\ \hline
\end{tabular}
\caption{OpenMP clause breakdown in \hpcorpusomp{}.
 \vspace{-0.7cm}}
\label{tab:omp_corpus}
%\vspace{-0.5cm}

\end{table}

\mypara{\comp{}-\textit{OMP} training.} We fine-tuned the pretrained \comp{} model to the OpenMP pragma generation task using the Huggingface \texttt{transformers} library on 2 NVIDIA V100 32GB GPUs at \texttt{fp32} precision. Fine-tuning used the AdamW optimizer with a linear warmup over the first 100 steps, followed by a linear decay over the remaining steps. Fine-tuning samples also had a maximum length of 2048 tokens, trained in 16 sample minibatches (8 per GPU). The fine-tuning was run for 4 epochs at the learning rate of 0.000016.

\mypara{Evaluation setup.} In the fine-tuning of \comp{} on the subcorpus of \hpcorpusomp{}, the task was set as a generation task in which the complete \texttt{for} loop is input to the model, and the expected generated output was the pragma and its associated variables. We then evaluated \comp{}-\textit{OMP} on a test set of \hpcorpusomp{}. %, both with the original \texttt{for} loops and their anonymized versions using the Local Semantics Elimination (LSE) pipeline~\cite{kadosh2023domain}.
%As a baseline, we conducted the finetuning of PolyCoder exclusively on the \hpcorpusomp{} subcorpus.
For comparison, we assessed GPT-3.5 turbo in a zero-shot manner. Specifically, we supplied a prompt instructing it to \textit{``Generate the optimal OpenMP pragma for the provided code''} when presented with the \texttt{for} loop.

% \ompifygen{} will be evaluated and compared against state-of-the-art source-to-source parallel compilers, such as autopar, and state-of-the-art internal parallelism compilers, such as icpc, in terms of both the performance of the parallel code and its correctness. Concretely,

\section{Experimental evaluation}

Our experimental evaluation is designed to answer the following research questions:
\begin{itemize}
    \item {\textbf{RQ1:}} How accurately can \ompifygen{} predict and suggest correct OpenMP pragma for a given \texttt{for} loop?
    \item {\textbf{RQ2:}} Do programs with \ompifygen{}-suggested pragmas compile and run correctly? If so, what is their runtime performance and scalability in a multi-core, shared-memory environment?
\end{itemize}

The first research question evaluates \ompifygen{} as an AI model; the second research question evaluates its ability to suggest correct and high-performance pragmas. This complementary approach validates \ompifygen{} effectiveness. Henceforth we will describe the experimental setup, the evaluation methodology in answering the two RQs, and finally the results.

\subsection{Experimental setup}

\mypara{HeCBench test dataset.} HeCBench is a novel heterogeneous computing benchmark suite with over 350 benchmarks, written in CUDA, SYCL, HIP, and OpenMP, spanning diverse domains, including machine learning, image processing, and simulation~\cite{jin2023benchmark}. More importantly, all the OpenMP benchmarks from HeCBench follow a uniform compilation recipe\footnote{Although HeCBench is a benchmark suite for heterogeneous environments, its OpenMP benchmarks could be compiled for CPU-only environment using \texttt{make DEVICE=cpu}.} and also contain a test harness that consists of test inputs and a procedure that compares the generated output with an expected reference output.

As HeCBench is an open-source benchmark, it is reasonable to believe that it could be part of \hpcorpus{} and as a result, part of \comp{}{}'s pre-training set. To eliminate this concern, we first ensured that HeCBench was not part of the pre-training or fine-tuning of \comp{}{}. The next step in synthesizing the test dataset of OpenMP \texttt{for} loops from HeCBench was to perform a basic sanity check of OpenMP benchmarks from HeCBench. Specifically, we attempted to compile them, run them with a timeout of 10 minutes (to choose a reasonable time on our test system), and check that all of the benchmarks passed the output correctness test. We found that 223 OpenMP benchmarks passed all the tests.

\begin{table}[!tbp]
\centering
\begin{tabular}{|l|r|}
\hline
\texttt{\textbf{Pragma (merged with \textit{target} form)}} & \texttt{\textbf{Total}} \\ \hline
\texttt{parallel} & 14 \\ \hline
\texttt{parallel for} & 311 \\ \hline
\texttt{parallel for private} & 3 \\ \hline
\texttt{parallel for reduction} & 27 \\ \hline
\texttt{parallel for simd} & 22 \\ \hline
\texttt{parallel for simd reduction} & 4 \\ \hline
\end{tabular}
\caption{Pragma breakdown in HeCBench test dataset. Although HeCBench is a benchmark suite for heterogeneous environments, its OpenMP benchmarks could be compiled for CPU-only environments using \texttt{make DEVICE=cpu}. Thus, for simplicity, we consolidated all matching pragmas starting with \texttt{\#pragma omp target teams distribute} and continuing with \texttt{parallel /+ for /+ private, reduction, simd}.
 \vspace{-0.7cm}}.
\label{tab:hecbench_pragma_breakdown}
%\vspace{-0.5cm}
\end{table}

\begin{table*}[!ht]
\centering
%\begin{footnotesize}
\begin{tabular}{c||r|r|r|r|r|r|r}
\hline
\textbf{Test setup} & \textbf{TP} & \textbf{FP} & \textbf{TN} & \textbf{FN} & \textbf{Precision} & \textbf{Recall} & \textbf{Accuracy} \\ \hline
\ompifygen{} accuracy with ground-truth label  & 311  & 127  & 262 & 70 & 71\% & \textbf{81\%} & \textbf{74\%} \\ \hline
AutoPar accuracy with ground-truth label  & 63  & 25  & 365 & 317 & 71\% & 17\% & 56\% \\ \hline
ICPC accuracy with ground-truth label & 95  & 11  & 285 & 379 & \textbf{90\%} & 25\% & 62\% \\ \Xhline{1.2pt}
\ompifygen{} accuracy with compile and run check & 407  & 31 & 262 & 70 & 92\% & \textbf{85\%} & \textbf{86\%} \\ \hline
AutoPar accuracy with compile and run check  & 24  & 25  & 365 & 356 & 49\% & 6\% & 50\% \\ \hline
ICPC accuracy with compile and run check  & 68  & 5  & 312 & 385 & \textbf{93\%} & 15\% & 49\% \\ \hline
\end{tabular}
%\end{footnotesize}
\caption{Result of accuracy test on HeCBench (The best scores are in bold).
\vspace{-0.7cm}}
\label{tab:accuracy_test}
%\vspace{-0.5cm}
\end{table*}

We then used these 223 OpenMP benchmarks to synthesize the HeCBench test dataset. Specifically, we obtained all the \texttt{for} loops from these benchmarks that had some OpenMP pragma\footnote{As \ompifygen{} does not support offloading-specific OpenMP pragmas such as \texttt{\#pragma omp target data}, we did not consider them for our selection.}. The breakdown of OpenMP pragmas from the collected loops is shown in \autoref{tab:hecbench_pragma_breakdown}. In all, the dataset contained 385 \texttt{for} loops having one of the OpenMP pragmas from the table. We balanced the dataset in terms of ``negative'' \texttt{for} loops --- loops that do not contain any OpenMP pragma --- by randomly choosing 385 such \texttt{for} loops from HeCBench's OpenMP benchmarks. In summary, our HeCBench dataset for the evaluation contained 770 \texttt{for} loops. In terms of OpenMP benchmarks, it covered 175 of 223 benchmarks.

\mypara{Competing tools.} We use AutoPar~\cite{liao2010semantic}, a static source-to-source-based automatic parallelization tool, and ICPC, Intel's auto-\linebreak parallelizing compiler, as baselines for this experiment. Specifically, because both AutoPar and ICPC operate on complete programs (as against a single \texttt{for} loop), we generate serial versions of HeCBench's OpenMP benchmarks by removing all OpenMP pragmas from them\footnote{For the pragma erasing, we used AutoParBench's PragmaRemover tool: \url{https://github.com/LLNL/AutoParBench/tree/master/tools/PragmaRemover}}. We then apply AutoPar and ICPC to parallelize those serial OpenMP benchmarks, attempt to compile generated parallelized OpenMP benchmarks, run with the test inputs, and compare their output with the expected output. Finally, we also measure the execution runtime for different values of \texttt{OMP\_NUM\_THREADS}. By comparing the performance of \ompifygen{} against these established tools (with AutoPar~\cite{quinlan2011rose} framework), we gain insights into the relative advantages and limitations of \ompifygen{}'s approach to parallel code generation. Note that we did not consider AI-based auto-parallelizing tools for our evaluation as both \ompify{} and \comp{} have performed extensive comparative evaluations of these tools; moreover, work by Nichols et al.~\cite{nichols2024can} has also performed comprehensive evaluations of the ability of popular LLMs for auto-parallelization.

\mypara{Versions.} For this evaluation, we used HeCBench open-source code with GitHub commit ID \texttt{927f02a}. We used AutoPar from ROSE-0.11.46.0.1, which is a state-of-the-art tool that is widely used for source-to-source transformation and is actively maintained. We used ICPC Intel Compiler 19.1.3.304 20200925. We used Intel(R) oneAPI DPC++/C++ Compiler 2022.0.0 for the compilation of HeCBench's OpenMP benchmarks. We ran compiled benchmarks on a dual-socket, 80-core, Intel Xeon Platinum 8380 CPU, running at 2.30GHz with hyper-threading enabled.

\subsection{Evaluation methodology}
\label{section:eval_rq2}

Our evaluation methodology is broken down into two sub-steps (\autoref{fig:teaser}, right):

%\begin{itemize}
    \mypara{(1) Accuracy test.} This test is designed to answer \textbf{RQ1}. In this test, we feed each loop out of 770 \texttt{for} loops to all three tools separately and compare the pragma generated by them or lack thereof with the ground-truth pragma (i.e., label). We use standard machine learning metrics (precision, recall, accuracy) to report the performance of every tool on this test.

    \mypara{(2) Compile \& run, scale and validation test.} This test is designed to answer \textbf{RQ2}. If the tool that we are evaluating generates some OpenMP pragma for the input \texttt{for} loop, then we insert that pragma in the OpenMP benchmark which contains that loop. In other words, we replace the original pragma of the loop with the inferred pragma. Then we attempt to compile that benchmark, execute it with the test input (that comes along with the benchmark), and compare the expected benchmark output with the actual output. We also measure the execution runtime for different values of \texttt{OMP\_NUM\_THREADS}, the environment variable that controls the number of threads in a \textit{parallel for} region. The execution runtime baseline is the HeCBench OpenMP benchmarks run in their default setting.
%\end{itemize}

\subsection{Results}

\begin{table*}[!ht]
\parbox{.6\linewidth}{
\centering
%\begin{footnotesize}
\begin{tabular}{c||r|r|r|r}
\hline
\textbf{Threads} & \textbf{C\&R pass} & \textbf{Compile fail} & \textbf{Run fail} & \textbf{Timeout} \\ \hline
1 & 706 (91.68\%) & 53 & 0 & 11 \\ \hline
4 & 710 (92.2\%) & 53 & 7 & 0 \\ \hline
8 & 710 (92.2\%) & 53 & 7 & 0 \\ \hline
16 & 707 (91.81\%) & 53 & 10 & 0 \\ \hline
\end{tabular}
%\end{footnotesize}
\caption{Compile and run (C\&R) test results for \ompifygen{} on 770 loops from HeCBench.\vspace{-0.7cm}}
\label{tab:compile_and_run_test_loops}
}
\hfill
\parbox{.35\linewidth}{
\centering
%\begin{footnotesize}
\begin{tabular}{c||r|r}
\hline
\textbf{Tool} & \textbf{Pass} & \textbf{Failed} \\ \hline
 \ompifygen{} & \textbf{130} & \textbf{45} \\ \hline
  AutoPar & 42 & 133 \\ \hline
  ICPC & 72 & 17 \\ \hline
\end{tabular}
\caption{Compile and run test results for 175 OpenMP benchmarks from HeCBench (The best scores are in bold). \vspace{-0.7cm}}
\label{tab:compile_and_run_test_dir}
%\end{footnotesize}
}
%\vspace{-0.5cm}
\end{table*}

\mypara{Results on accuracy test.} \autoref{tab:accuracy_test} shows the performance of \ompifygen{} on HeCBench test set of 770 loops. \ompifygen{} accurately predicted the pragma in 74\% of the test loops (details in the first row of the table). %Specifically, after considering 770 loops to evaluate the pragma prediction performance of \ompifygen{}, we realized that it was reporting a decent accuracy of 74\% (details in the first row of the table).

%After careful analysis, we found that 
\ompifygen{} appeared to make many false positive predictions, i.e. it predicted parallelization when those loops originally had no OpenMP pragma (i.e., ``negative'' \texttt{for} loops).  
%was predicting the possibility of parallelizing several \texttt{for} loops that originally had no OpenMP pragma (i.e., ``negative'' for loops). 
To confirm if \ompifygen{} suggested loops can indeed be parallelized using the pragma it was suggesting, we decided to subject all 438 positively predicted loops (i.e., 311 TP + 127 FP) to the compile and run test. Specifically, we compiled OpenMP benchmarks by inserting \ompifygen{} suggested pragmas for a given loop, ran those benchmarks with their test inputs, and verified if their output matched the expected output.

To our surprise, we found that 96 of 127 false positives could actually be compiled and run -- meaning 96 loops without OpenMP pragmas in the ground truth could, in fact, be parallelized using \ompifygen{}-suggested OpenMP pragmas. The remaining 31 false positives were indeed loops that could not be parallelized. In this particular case, we consider compile and run tests to have higher credibility (as a test oracle) over the original loop label to determine the potential of parallelizing a loop. This is because a loop may not be parallelized for several reasons such as the programmer missing out on the opportunity, potential performance improvement with the parallelization, etc. After taking the compile and run check into account, \ompifygen{} achieved 86\% accuracy, 92\% precision, and 85\% recall (\autoref{tab:accuracy_test}).

We also evaluated AutoPar and  ICPC using the same methodology (\autoref{tab:accuracy_test}). Higher precision values of ICPC point to the conservative nature of ICPC --- it has low false positives. On the other hand, its low recall rate (high false negatives) indicates that it missed many parallelization opportunities. AutoPar shows a similar behavior trait as ICPC, though its precision and recall scores are lower.

\begin{tcolorbox}[colback=blue!5!white,colframe=black]
  \textbf{RQ1: Overall, we believe our accuracy test results of the three tools on OpenMP benchmarks from HeCBench reveals that: (1) \ompifygen{} clearly outperforms AutoPar and ICPC, deterministic tools for parallelization, in terms of accurately determining the parallelization potential a loop as well as in determining the actual pragma and its details, and (2) \ompifygen{} demonstrates a good balance of precision and recall scores, arguably a sweet-spot in terms of prediction performance of an auto-parallelization tool.}
\end{tcolorbox}

\mypara{Results on compile and run test.} We performed the compile and run test on all 770 loops from \ompifygen{}'s accuracy test results and found that 717 of those passed the compilation test, while 706-710 of those 717 could pass the output verification test for different settings of \texttt{OMP\_NUM\_THREADS}. \autoref{tab:compile_and_run_test_loops} shows the detailed results. In summary, for \ompifygen{}, around 92\% of the 770 loops successfully passed the compilation as well as the output verification test.

In order to account for different operating modes of \ompifygen{} and AutoPar (i.e., parallelizing a loop individually vs parallelizing the whole program), we also report the performance of both at the level of OpenMP benchmarks (as against individual loop) in HeCBench (\autoref{tab:compile_and_run_test_dir}). Specifically, out of a total of 175 OMP benchmarks covered by 770 loops, \ompifygen{} could successfully parallelize all the loops from 130 benchmarks that also passed compile and run test (output verification). To be precise, loops belonging to the remaining 45 benchmarks either could not be parallelized by \ompifygen{} or the parallelized loops failed compilation/run check. AutoPar, on the other hand, could parallelize a total of 109 benchmarks, with 63 of those passing the compilation test and 42 passing the output verification test. \autoref{tab:compile_and_run_test_dir} shows the overall results for 175 OpenMP benchmarks from HeCBench.

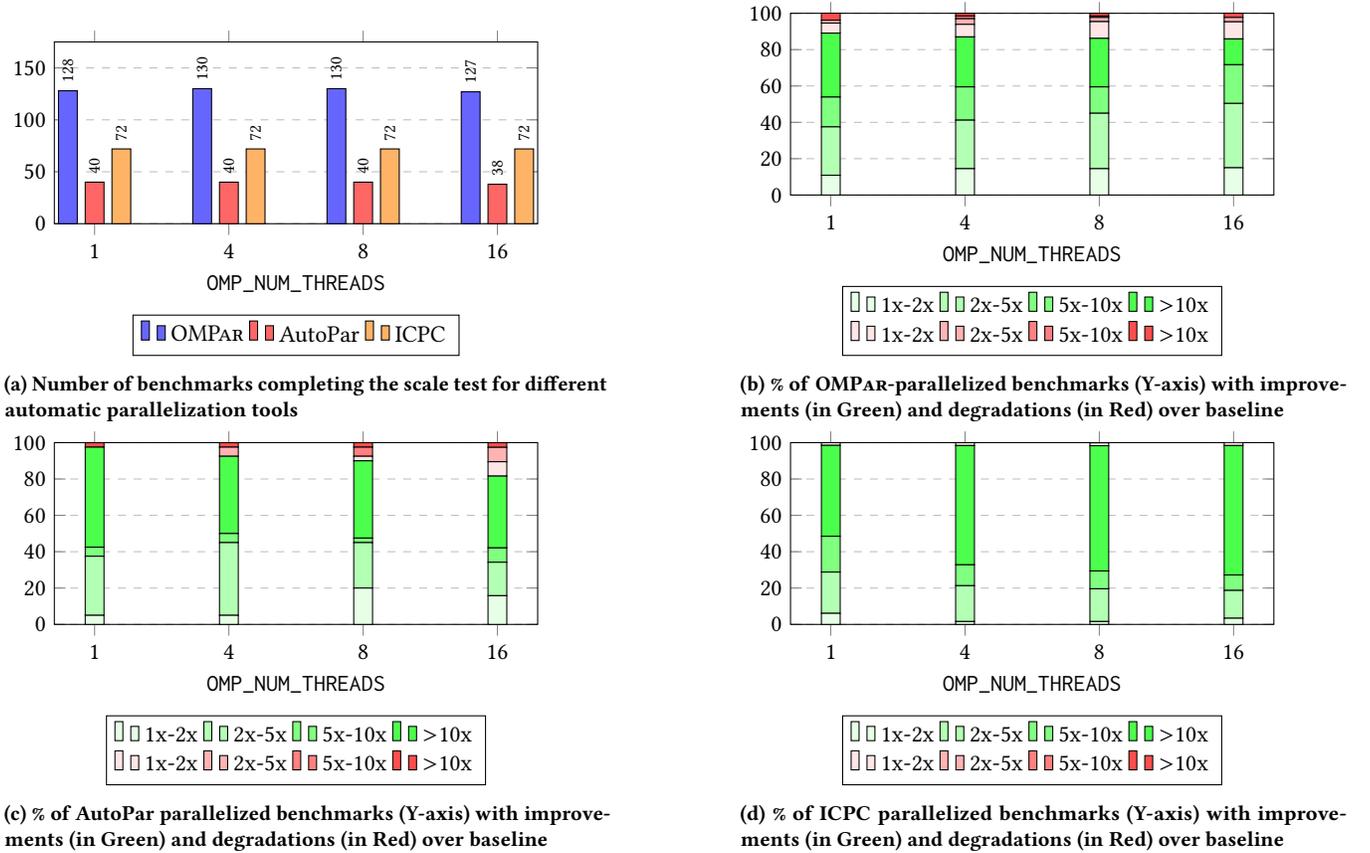
\begin{figure*}[!ht]
    %\centering
    %\begin{minipage}{0.75\textwidth}
        %\centering
        %\hspace{0.8cm}
        \begin{subfigure}[b]{0.45\linewidth}
        \begin{tikzpicture}
        \centering[height=6cm]
            \begin{axis}[
                legend columns=3,
                ybar,
                bar width=0.25cm, 
                symbolic x coords={1, 4, 8, 16},
                xlabel={\texttt{OMP\_NUM\_THREADS}},
                xtick=data,
                ymin=0,
                ymax=175,
                height=4cm,
                width=\linewidth,
                nodes near coords,
                every node near coord/.append style={rotate=90, anchor=west, font=\scriptsize},
                ybar=3pt, 
                legend style={at={(0.5,-0.5)},anchor=north,legend columns=-1},
                grid style=dashed,
                ymajorgrids=true,
            ]
            \addplot[fill=blue!60] coordinates {
                (1, 128)
                (4, 130)
                (8, 130)
                (16, 127)
            };
            \addplot[fill=red!60] coordinates {
                (1, 40)
                (4, 40)
                (8, 40)
                (16, 38)
            };
            \addplot[fill=orange!60] coordinates {
                (1, 72)
                (4, 72)
                (8, 72)
                (16, 72)
            };
            \legend{\ompifygen{}, AutoPar, ICPC}
            \end{axis}
        \end{tikzpicture}
        \caption{Number of benchmarks completing the scale test for different automatic parallelization tools}
        \label{fig:scale_test_benchmark_number}
    \end{subfigure}
    %\hspace{1cm}
    \hfill
    %
    %\centering
    \begin{subfigure}[b]{0.45\linewidth}
        \begin{tikzpicture}
        \centering[height=6cm]
            \begin{axis}[
                legend columns=4,
                ybar stacked,
                bar width=0.25cm, 
                symbolic x coords={1, 4, 8, 16},
                xlabel={\texttt{OMP\_NUM\_THREADS}},
                xtick=data,
                ymin=0,
                ymax=100,
                height=4cm,
                width=\linewidth,
                %nodes near coords,
                every node near coord/.append style={rotate=90, anchor=west, font=\scriptsize},
                ybar=3pt, 
                legend style={at={(0.5,-0.5)},anchor=north,legend columns=-1},
                grid style=dashed,
                ymajorgrids=true,
                bar shift=0pt
            ]
        %\begin{axis}[,hide axis]
        \addplot[fill=green!10] coordinates {
            % improvement 1x-2x
           (1, 10.9) (4, 14.5) (8, 14.5) (16, 15)
        };
        \addplot[fill=green!30] coordinates {
            % imrovement 2x-5x
           (1, 26.6) (4, 26.7) (8, 30.5) (16, 35.4)
        };
        \addplot[fill=green!50] coordinates {
            % improvement 5x-10x
           (1, 16.4) (4, 18.3) (8, 14.5) (16, 21.3)
        };
        \addplot[fill=green!70] coordinates {
            % improvement > 10x
           (1, 35.2) (4, 27.5) (8, 26.7) (16, 14.2)
        };
        \addplot[fill=red!10] coordinates {
            % degradation 1x-2x
           (1, 5.5) (4, 6.9) (8, 9.2) (16, 9.4)
        };
        \addplot[fill=red!30] coordinates {
            % degradation 2x-5x
           (1, 1.6) (4, 3.1) (8, 2.3) (16, 2.4)
        };
        \addplot[fill=red!50] coordinates {
            % degradation 5x-10x
           (1, 0) (4, 1.5) (8, 0.8) (16, 0)
        };
        \addplot[fill=red!70] coordinates {
            % degradation > 10x
           (1, 3.9) (4, 1.5) (8, 1.5) (16, 2.4)
        };
        \legend{1x-2x, 2x-5x, 5x-10x, $>$10x, 1x-2x, 2x-5x, 5x-10x, $>$10x}
        \end{axis}
        \end{tikzpicture}
        \caption{\% of \ompifygen{}-parallelized benchmarks (Y-axis)  with improvements (in Green) and degradations (in Red) over baseline}
        \label{fig:scale_test_percentage_ompifygen}
    \end{subfigure}
    %\hspace{1cm}

    %
    \begin{subfigure}[b]{0.45\linewidth}
        \begin{tikzpicture}
        \centering[height=6cm]
            \begin{axis}[
                legend columns=4,
                ybar stacked,
                bar width=0.25cm, 
                symbolic x coords={1, 4, 8, 16},
                xlabel={\texttt{OMP\_NUM\_THREADS}},
                xtick=data,
                ymin=0,
                ymax=100,
                height=4cm,
                width=\linewidth,
                %nodes near coords,
                every node near coord/.append style={rotate=90, anchor=west, font=\large},
                ybar=3pt, 
                legend style={at={(0.5,-0.5)},anchor=north,legend columns=-1},
                grid style=dashed,
                ymajorgrids=true,
                bar shift=0pt
            ]
        \addplot[fill=green!10] coordinates {
            % improvement 1x-2x
           (1, 5) (4, 5) (8, 20) (16, 15.8)
        };
        \addplot[fill=green!30] coordinates {
            % improvement 2x-5x
           (1, 32.5) (4, 40) (8, 25) (16, 18.4)
        };
        \addplot[fill=green!50] coordinates {
            % improvement 5x-10x
           (1, 5) (4, 5) (8, 2.5) (16, 7.9)
        };
        \addplot[fill=green!70] coordinates {
            % improvement > 10x
           (1, 55) (4, 42.5) (8, 42.5) (16, 39.5)
        };
        \addplot[fill=red!10] coordinates {
            % degradation 1x-2x
           (1, 0) (4, 0) (8, 2.5) (16, 7.9)
        };
        \addplot[fill=red!30] coordinates {
            % degradation 2x-5x
           (1, 0) (4, 5) (8, 0) (16, 7.9)
        };
        \addplot[fill=red!50] coordinates {
            % degradation 5x-10x
           (1, 0) (4, 0) (8, 5) (16, 0)
        };
        \addplot[fill=red!70] coordinates {
            % degradation > 10x
           (1, 2.5) (4, 2.5) (8, 2.5) (16, 2.6)
        };
        \legend{1x-2x, 2x-5x, 5x-10x, $>$10x, 1x-2x, 2x-5x, 5x-10x, $>$10x}
        \end{axis}
        \end{tikzpicture}
        \caption{\% of AutoPar parallelized benchmarks (Y-axis) with improvements (in Green) and degradations (in Red) over baseline}
        \label{fig:scale_test_percentage_autopar}
    \end{subfigure}
    \hfill
    \begin{subfigure}[b]{0.45\linewidth}
        \begin{tikzpicture}
        \centering[height=6cm]
            \begin{axis}[
                legend columns=4,
                ybar stacked,
                bar width=0.25cm, 
                symbolic x coords={1, 4, 8, 16},
                xlabel={\texttt{OMP\_NUM\_THREADS}},
                xtick=data,
                ymin=0,
                ymax=100,
                height=4cm,
                width=\linewidth,
                %nodes near coords,
                every node near coord/.append style={rotate=90, anchor=west, font=\scriptsize},
                ybar=3pt, 
                legend style={at={(0.5,-0.5)},anchor=north,legend columns=-1},
                grid style=dashed,
                ymajorgrids=true,
                bar shift=0pt
            ]
        \addplot[fill=green!10] coordinates {
            % improvement 1x-2x
           (1, 6.1) (4, 1.6) (8, 1.6) (16, 3.4)
        };
        \addplot[fill=green!30] coordinates {
            % improvement 2x-5x
           (1, 22.7) (4, 19.7) (8, 18) (16, 15.3)
        };
        \addplot[fill=green!50] coordinates {
            % improvement 5x-10x
           (1, 19.7) (4, 11.5) (8, 9.8) (16, 8.5)
        };
        \addplot[fill=green!70] coordinates {
            % improvement > 10x
           (1, 50) (4, 65.6) (8, 68.9) (16, 71.2)
        };
        \addplot[fill=red!10] coordinates {
            % degradation 1x-2x
           (1, 1.5) (4, 1.6) (8, 1.6) (16, 1.7)
        };
        \addplot[fill=red!30] coordinates {
            % degradation 2x-5x
           (1, 0) (4, 0) (8, 0) (16, 0)
        };
        \addplot[fill=red!50] coordinates {
            % degradation 5x-10x
           (1, 0) (4, 0) (8, 0) (16, 0)
        };
        \addplot[fill=red!70] coordinates {
            % degradation > 10x
           (1, 0) (4, 0) (8, 0) (16, 0)
        };
        \legend{1x-2x, 2x-5x, 5x-10x, $>$10x, 1x-2x, 2x-5x, 5x-10x, $>$10x}
        \end{axis}
        \end{tikzpicture}
        \caption{\% of ICPC parallelized benchmarks (Y-axis) with improvements (in Green) and degradations (in Red) over baseline}
        \label{fig:scale_test_percentage_icpc}
    \end{subfigure}
    %
    % \hfill
    % %
    % \begin{subfigure}[b]{0.33\textwidth}
    %     \centering
    %     \includegraphics[width=\linewidth]{plots/ompar_vs_icpc_and_autopar.pdf}
    %     \caption{Performance comparison of \ompifygen{}-, ICPC- and AutoPar-parallelized benchmarks \niranjan{Will fix this}}
    %     \label{fig:ompar_icpc_autopar}
    % \end{subfigure}
    % %
    \caption{Scale test performance of \ompifygen{}-, AutoPar-, and ICPC-parallelized OpenMP benchmarks from HeCBench.}
    \label{fig:scale_test}
\end{figure*}

\mypara{Results on scale test.} The objective of the scale test is to ensure that the pragmas suggested by \ompifygen{} do not degrade the performance of the OpenMP benchmarks from HeCBench. Towards that end, we ran all 130 OpenMP benchmarks that were parallelized using \ompifygen{} on a datacenter-class, dual-socket, 40 cores per socket, Intel Xeon Platinum 8380 processor on the 4 different values of \texttt{OMP\_NUM\_THREADS} that were used in the compile and run test. We also ran those benchmarks in their default settings and without setting any specific value of \texttt{OMP\_NUM\_THREADS} as a baseline. %We repeated the same experiment for AutoPar, but for the 42 benchmarks, it could parallelize and pass the compile and run test.
We repeated the same experiment for AutoPar and ICPC, but only for the benchmarks that passed the compile and run test (42 and 72, respectively).

\autoref{fig:scale_test} shows the overall performance of \ompifygen{}, AutoPar, and ICPC on the scale test. Specifically, \autoref{fig:scale_test_benchmark_number} shows the number of benchmarks that were parallelized by both the tools for different values of \texttt{OMP\_NUM\_THREADS}. Overall, the figure shows that the tools can parallelize most of the benchmarks that also scale well (do not timeout) for different numbers of threads (except 3 for \ompifygen{} with 1 thread and 16 threads and 2 for AutoPar with 16 threads). \autoref{fig:scale_test_percentage_ompifygen}--\autoref{fig:scale_test_percentage_icpc} provide further details in terms of performance improvements/degradations of the benchmarks over baseline.
%for \ompifygen{} and AutoPar, respectively.
In particular, we categorize the performance of every benchmark over baseline into improvement (shown in green) or degradation (shown in red) of different magnitudes (1x-2x, 2x-5x, 5x-10x, and $>$10x, most saturated is largest magnitude). Since %\ompifygen{} and AutoPar 
the three tools parallelize a different number of benchmarks, the improvements or degradations are shown as a percentage of the number of tests for easy comparison across tools. %we calculate percentages of the benchmarks (on the Y axis) belonging to these 8 categories to allow easier comparison. 
Overall, the results show that (1) the percentage of benchmarks degrading over baseline is considerably less than those showing the improvement; (2) although AutoPar is able to achieve $>$10x improvement for more \% of benchmarks than \ompifygen{}, \ompifygen{} is showing similar performance, even though it is parallelizing almost 3X more benchmarks than AutoPar; (3) the conservative nature of ICPC results in very few performance degradations, although it is parallelizing about half the number of benchmarks as \ompifygen{}.

\begin{tcolorbox}[colback=blue!5!white,colframe=black]
  \textbf{RQ2: Overall, we believe our evaluations of the three tools on compile, run, and scale tests reveal two key findings: (1) More than 90\% of the loops parallelized using \ompifygen{} pass the compile and run test, suggesting that \ompifygen{} understands the syntax as well as the semantics of OpenMP pragmas and the loops that they apply to, and (2) \ompifygen{} parallelized OpenMP benchmarks show orders of magnitude performance improvement over baseline, and even after parallelizing more number of benchmarks, \ompifygen{} parallelized benchmarks perform similar (if not better) to AutoPar parallelized benchmarks.}
\end{tcolorbox}

\subsection{Result analysis: What were some of AutoPar failures?}

Given that AutoPar is a rule-based, state-of-the-art source-to-source transformation tool for automatic parallelization, it is natural to expect it to parallelize most, if not all, of the possible \texttt{for} loops. Our experimental evaluation on HeCBench revealed otherwise. Hence, we decided to investigate some of the cases where AutoPar failed to parallelize or incorrectly parallelized a loop. Below we present the most frequent failures that we observed in our experiments.

\subsubsection{\textbf{Cases of incorrect parallelization}}

AutoPar parallelized code failed to compile, prompting further analysis for examination. Upon investigation, we found that AutoPar had performed several erroneous code transformations.

\mypara{(1) Applies parallelization (i.e., adds \texttt{\#pragma}) to non-canonical loops.} Section 2.6 of OpenMP-v4.5~\cite{openmp2015arb} specifies a particular form of loops (called \emph{canonical form}) that can be parallelized using OpenMP. All other loop forms are then non-canonical forms.

\begin{lstlisting}[]
// AutoPar suggested OpenMP pragma
#pragma omp parallel for private (i)
for (i = 0;
     ((long) i) <= dim_cpu.space_elem - 1;
     i = i + 1) {
  rv_cpu[i].v = ((rand() % 10 + 1) / 10.0);
  ...}
\end{lstlisting}

We found that AutoPar suggested OpenMP parallel pragma for non-canonical loops also. For example, a loop shown above (obtained from HeCBench's \texttt{lavaMD-omp}) contains a typecasting expression involving the loop iteration variable, and such an expression is considered non-canonical by OpenMP. In spite of this, AutoPar parallelized the particular loop, thus leading to a compiler error.

\mypara{(2) Applies parallelization to loops with \texttt{return} statements.} OpenMP specifications do not allow the parallelization of loops containing a \texttt{return} statement in its body. This constraint ensures that all threads synchronize before exiting the code. In spite of this restriction, AutoPar suggested OpenMP parallelization pragma for such a loop as shown in the example obtained from HeCBench's \texttt{grep-omp} benchmark.

\begin{lstlisting}[]
inline int ispmatch(::List *l) {
  int i;
  // AutoPar suggested OpenMP pragma
  #pragma omp parallel for private (i)
  for (i = 0; i <= l->n-1; i += 1) {
    if (l->s[i]->c == Match)
      return 1;}
  return 0;}
\end{lstlisting}

\mypara{(3) Applies privatization and reduction to the same variable in OpenMP parallelization pragma.} Privatization and reduction of the same variable are not allowed in OpenMP because \texttt{private} and \texttt{reduction} are conflicting directives for the same variable. In spite of this rule, AutoPar suggested OpenMP pragma for a loop (example below from \texttt{page-rank-omp} benchmark of HeCBench) that violated the rule.

\begin{lstlisting}[]
// AutoPar suggested OpenMP pragma
#pragma omp parallel for \
  private (nb_links,i,j) reduction (+:nb_links)
for (i = 0; i <= n - 1; i += 1) {
  #pragma omp parallel for \
    private (j) reduction (+:nb_links)
  for (j = 0; j <= n - 1; j += 1) {
    nb_links += pages[i * n + j];}}
\end{lstlisting}

\subsubsection{\textbf{Cases of missed parallelization opportunities}}

Performance analysis of AutoPar-parallelized programs revealed that AutoPar had missed parallelization opportunities in several HeCBench benchmarks.

One commonly occurring theme among the loops that AutoPar missed was that many of them contained function calls. For instance, in the code shown below (obtained from \texttt{ace-omp} benchmark) AutoPar did not parallelize the loop as the loop body contains multiple function calls. Specifically, for loops containing function calls, AutoPar assumes that the function call might modify shared data, or the callee may contain loops, or the call may be data-dependent. AutoPar does not perform inter-procedural analysis or optimizations, such as function inlining, which could have alleviated this limitation. AutoPar missed parallelization opportunities at multiple places within the same benchmark for this exact same reason.

\begin{lstlisting}[]
for (int ix = 0; ix <= 99; ix += 1) {
 for (int iy = 0; iy <= 99; iy += 1) {
   for (int iz = 0; iz <= 99; iz += 1) {
    if (ix < 100 - 1 && iy < 100 - 1 &&
        iz < 100 - 1 && ix > 0 && iy > 0 &&
        iz > 0) {
      double px =
        GradientX(phi,dx,dy,dz,ix,iy,iz);
      double py =
        GradientY(phi,dx,dy,dz,ix,iy,iz);
      double pz =
        GradientZ(phi,dx,dy,dz,ix,iy,iz);
      double sqGphi = px * px + py * py +
                      pz * pz;
      double c = 16.0 * W0 * epsilon;
      double w = Wn(px,py,pz,epsilon,W0);
      double w2 = w * w;
      Fx[ix][iy][iz] = w2 * px +
        sqGphi * w * c * dFunc(px, py, pz);
      Fy[ix][iy][iz] = w2 * py +
        sqGphi * w * c * dFunc(py, pz, px);
      Fz[ix][iy][iz] = w2 * pz +
        sqGphi * w * c * dFunc(pz, px, py);
    } else {
      Fx[ix][iy][iz] = 0.0;
      Fy[ix][iy][iz] = 0.0;
      Fz[ix][iy][iz] = 0.0;}}}} 
\end{lstlisting}

In summary, we found that even AutoPar, which is a state-of-the-art source-to-source transformation tool based on a formal, rule-based approach, either missed parallelization opportunities or suggested incorrect pragmas. We believe this behavior can be attributed to missing newer rules (e.g., limited support for OpenMP-4.5) or conservative assumptions. Our experimental results reveal that AI-based approaches, such as ours, can mitigate these limitations, by learning from vast amounts of code and identifying parallelization patterns that rule-based systems may overlook. 

\subsection{Evaluating \ompifygen{} on ParEval dataset}

\begin{figure*}[!htb]
\centering
\begin{minipage}[b]{0.45\textwidth}
%\begin{footnotesize}
\begin{tabular}{p{2cm}||r|r|r|r|p{0.5cm}|p{0.5cm}|p{0.5cm}}
\hline
\textbf{Test setup} & \textbf{TP} & \textbf{FP} & \textbf{TN} & \textbf{FN} & \textbf{P} & \textbf{R} & \textbf{Acc} \\ \hline
With ground-truth label  & 37  & 18  & 0 & 0 & 67\% & 100\% & 67\% \\ \hline
With compile and run check & 48  & 7 & 0 & 0 & 87\% & 100\% & 87\% \\ \hline
\end{tabular}
%\end{footnotesize}
\caption{\ompifygen{}'s performance on the accuracy test on the ParEval OpenMP dataset. (TP=True Positive, FP=False Positive, TN=True Negative, FN=False Negative, P=Precision, R=Recall, Acc=Accuracy)}
\label{tab:accuracy_test_pareval}
\end{minipage}
\hfill
\begin{minipage}[b]{0.45\textwidth}
\centering
%\begin{figure}{\textwidth}
    \begin{tikzpicture}
    \begin{axis}[
        axis lines = left,
        legend columns=3,
        ymin=0,
        ymax=150,
        bar width=0.15cm,
        xmin=1,
        xmax=48,
        ylabel={Speedup over nthreads=1 (baseline)},
        xlabel={Loop number},
        symbolic x coords={1,2,3,4,5,6,7,8,9,10,11,12,13,14,15,16,17,18,19,20,21,22,23,24,25,26,27,28,29,30,31,32,33,34,35,36,37,38,39,40,41,42,43,44,45,46,47,48},
        xticklabels={1,,,4,,,,8,,,,12,,,,16,,,,20,,,,24,,,,28,,,,32,,,,36,,,,40,,,,44,,,,48},
        xtick=data,
        width=\textwidth, % Adjusted width
        height=5cm,
        legend style={at={(0.5,-0.3)}, anchor=north, legend columns=-1},
        legend={},
        %cycle list/Set2,
        every node near coord/.append style={opacity=1},
    ]
    \addplot+[line width=0.5pt, blue!60, mark options={fill=blue!60}] plot coordinates {(1,0.07) (2,0.09) (3,0.17) (4,0.75) (5,0.79) (6,0.80) (7,1.22) (8,1.27) (9,1.29) (10,1.35) (11,1.38) (12,1.53) (13,1.53) (14,1.59) (15,1.62) (16,1.64) (17,1.65) (18,1.86) (19,1.93) (20,2.04) (21,2.37) (22,2.43) (23,2.45) (24,2.51) (25,2.79) (26,2.79) (27,3.13) (28,3.17) (29,3.18) (30,3.19) (31,3.29) (32,3.35) (33,3.82) (34,3.83) (35,3.88) (36,4.24) (37,4.83) (38,5.34) (39,5.36) (40,5.37) (41,5.85) (42,7.36) (43,7.62) (44,9.01) (45,10.38) (46,11.21) (47,13.04) (48,15.45)};
    \addplot+[line width=0.5pt, red!60, mark options={fill=red!60}] plot coordinates {(1,0.09) (2,0.09) (3,0.17) (4,0.90) (5,0.90) (6,0.89) (7,1.40) (8,1.41) (9,1.69) (10,1.71) (11,1.68) (12,1.85) (13,1.76) (14,1.91) (15,1.88) (16,1.85) (17,1.96) (18,2.10) (19,1.68) (20,2.34) (21,2.50) (22,2.67) (23,2.72) (24,2.66) (25,4.79) (26,6.43) (27,2.93) (28,2.64) (29,4.08) (30,4.22) (31,4.32) (32,4.35) (33,5.94) (34,5.67) (35,7.12) (36,7.12) (37,10.50) (38,9.16) (39,12.35) (40,8.16) (41,13.78) (42,23.63) (43,21.23) (44,26.29) (45,18.98) (46,22.26) (47,30.91) (48,41.12)};
    \addplot+[line width=0.5pt, orange!60, mark options={fill=orange!60}] plot coordinates {(1,0.09) (2,0.09) (3,0.33) (4,0.98) (5,1.01) (6,0.99) (7,1.69) (8,1.75) (9,2.24) (10,2.29) (11,2.26) (12,2.47) (13,2.42) (14,2.55) (15,2.41) (16,2.16) (17,2.64) (18,2.60) (19,2.24) (20,2.41) (21,2.71) (22,3.13) (23,3.22) (24,2.96) (25,8.93) (26,8.93) (27,3.46) (28,2.91) (29,5.83) (30,6.41) (31,6.58) (32,6.48) (33,9.94) (34,9.50) (35,10.18) (36,13.41) (37,20.00) (38,13.61) (39,38.47) (40,14.19) (41,36.22) (42,64.92) (43,62.29) (44,71.25) (45,40.38) (46,46.56) (47,102.44) (48,129.01)};
    \legend{nthreads=4, nthreads=8, nthreads=16}
    \end{axis}
    \end{tikzpicture}
    \caption{Performance of \ompifygen{} on the scale test with 48 True Positives from ParEval OpenMP dataset}
    \label{fig:scale_test_pareval}
%\end{figure}
\end{minipage}
\end{figure*}

Since we presented the performance of AutoPar and ICPC on the ParEval dataset in the motivation section (\autoref{section:motivation}), we decided to evaluate \ompifygen{} on the ParEval dataset as well. Notice that ParEval is a much smaller dataset than HeCBench, and \ompifygen{} has outperformed both AutoPar and ICPC on HeCBench. But we are presenting ParEval results for the sake of completeness and curious readers. As the performance of AutoPar and ICPC on ParEval was poor (parallelized only 2 benchmarks out of 6), we did not evaluate them on accuracy and scale tests.

\mypara{ParEval OpenMP loop dataset.} As we are only interested in obtaining OpenMP programs, we fed ParEval's 60 coding problems to GPT-3.5 and collected the output OpenMP programs. ParEval also comes with a set of compilation recipes and test inputs to evaluate if LLM-generated programs are syntactically- and semantically-correct. Using these recipes and inputs, we found that 32 programs of those 60 passed the compilation and runtime check. We then used these 32 OpenMP programs to extract 55 \texttt{for} loops, 37 of which had OpenMP \texttt{parallel for} pragma (``positive'' loops) while the remaining 18 had no OpenMP pragma (``negative'' loops). We call the dataset of 55 loops as ParEval OpenMP dataset.

\mypara{Evaluation methodology.} We used the dataset of 55 \texttt{for} loops to evaluate \ompifygen{} on the accuracy test and the compile, run, and scale test (discussed in \autoref{section:eval_rq2}).

\mypara{Results on accuracy test.} \autoref{tab:accuracy_test_pareval} shows the performance of \ompifygen{} on the accuracy test when subjected to 55 loops from the ParEval OpenMP dataset. As can be seen in the first row of the table, we found that the accuracy was relatively low (67\%). The reason is also obvious from the table --- we found that \ompifygen{} was suggesting that all of 18 ``negative'' \texttt{for} loops can be parallelized using OpenMP pragma. To confirm if this suggestion was indeed correct, we decided to subject \ompifygen{} to compile and run the test. Similar to the observations made from the evaluation on HeCBench, we found that 11 of those 18 ``negative'' loops can indeed be parallelized using suggested OpenMP pragma, and more importantly, the suggested pragma and the loop when inserted in the program can compile correctly and also pass output verification check. In other words, 11 of those 18 false positives are actually true positives. Consequently, the accuracy of \ompifygen{} improved to 87\% (as shown in the second row of the table).

\mypara{Results of scale test.} \autoref{fig:scale_test_pareval} shows the performance of \ompifygen{} on the scale test with ParEval OpenMP loop dataset. For the scale test, we insert the loop along with its \ompifygen{} suggested pragma in the original pragma and run the program with its default inputs. More importantly, we vary the number of threads \texttt{OMP\_NUM\_THREADS} from 1, 4, 8, and 16 while running those programs. The X-axis in the figure represents every loop from the dataset, while the Y-axis shows the speedup obtained by its corresponding program against the baseline of running the same program with 1 thread. Overall, as can be seen, as the number of threads is increased from 4 to 16, the performance of the programs goes up as well -- indicating that \ompifygen{} suggested pragmas scale well with different number of threads.

\section{Conclusions}
\label{conc}
In this paper, we introduced \ompifygen{}, an AI-driven source-to-source parallelization tool that leverages the strengths of Large Language Models (LLMs) to enhance automatic parallelization. By integrating the capabilities of \ompify{} for assessing loop parallelization feasibility and \comp{}-\textit{OMP} for generating precise OpenMP pragmas, \ompifygen{} significantly outperforms traditional auto-parallelization tools such as AutoPar and ICPC. Our extensive evaluation using the HeCBench and ParEval benchmarks demonstrates notable improvements in both accuracy and runtime performance. These results underscore the transformative potential of AI-driven approaches in optimizing parallel computing tasks. As we continue to refine and expand \ompifygen{}, we anticipate even broader applications and more robust performance in diverse computational environments.

\section{Future Work}
\label{summ}

While \ompifygen{} has demonstrated significant advancements in automatic parallelization, there are several promising directions for future research and development. These directions aim to extend the capabilities of \ompifygen{} and address its current limitations, paving the way for more versatile and powerful parallelization tools.

\textbf{Extending to GPU Architectures}: One immediate extension of \ompifygen{} is to broaden its applicability to GPU architectures. Current implementations are focused on multi-core CPU environments using OpenMP pragmas. To facilitate GPU parallelization, we plan to train the models with OpenMP offloading pragmas. This will involve curating a comprehensive dataset of code snippets that leverage GPU offloading and fine-tuning our models accordingly. As our method is already using HeCBench, we can seamlessly transition the evaluation to GPUs.

\textbf{Leveraging Large Language Models}: In this work, we utilized relatively small-scale models for the tasks of loop parallelization determination and pragma generation. However, the rapid advancement of LLMs such as GPT-4, Gemini, and CodeLLaMA presents an opportunity to significantly enhance the performance and accuracy of \ompifygen{}. Future work will involve experimenting with these state-of-the-art LLMs, fine-tuning them on our specialized datasets, and evaluating their effectiveness in the context of automatic parallelization. The superior contextual understanding and code generation capabilities of these larger models hold the potential to further elevate the performance of \ompifygen{}.

\textbf{Incorporating Diverse Code Representations}: A notable limitation of our current approach is that the AI models were trained and evaluated using plain source code. In contrast, traditional source-to-source parallel compilers utilize a variety of code representations such as ASTs, DFGs, and Intermediate Representations (IRs) to analyze and determine parallelization opportunities. To make a fairer comparison and potentially improve the performance of our AI models, future work will involve re-training the models using these diverse code representations. By integrating multiple modalities of code analysis, we expect to enhance the models' ability to accurately identify parallelization opportunities and generate appropriate pragmas.

\textbf{Comprehensive Benchmarking and Evaluation}: While our current evaluation focused on specific benchmarks such as HeCBench and ParEval, future work will include a more extensive benchmarking effort across a wider range of applications and domains. This will not only validate the robustness and scalability of \ompifygen{} but also provide deeper insights into its performance characteristics in various real-world scenarios. Additionally, we plan to conduct detailed performance profiling to identify and address any bottlenecks in the current implementation.

\textbf{Enhancing User Experience and Integration}: Finally, improving the usability and integration of \ompifygen{} within existing development workflows is a key area for future work. This includes developing intuitive user interfaces and providing seamless integration with popular Integrated Development Environments (IDEs). By making \ompifygen{} more accessible and user-friendly, we aim to empower a broader range of developers to harness the benefits of automatic parallelization.

% \clearpage
% \begin{acks}
% This research was supported by the Israeli Council for Higher Education (CHE) via the Data Science Research Center, Ben-Gurion University of the Negev, Israel; Intel Corporation (oneAPI CoE program); and the Lynn and William Frankel Center for Computer Science. Computational support was provided by HPE HPC \& AI Cloud \cite{breckenridge}, Intel Developer Cloud~\cite{intel-cloud}, and the NegevHPC project~\cite{negevhpc}.
% \end{acks}

%\clearpage
%\bibliographystyle{IEEEtran}
%\bibliography{IEEEabrv, references} % Specify your bibliography file

\begin{acks}
This research was supported by the Israeli Council for Higher Education (CHE) via the Data Science Research Center, Ben-Gurion University of the Negev, Israel; Intel Corporation (oneAPI CoE program); Pazy Foundation; and the Lynn and William Frankel Center for Computer Science. Computational support was provided by \textcolor{blue}{\href{https://console.breckenridge.cloud/}{HPE HPC \& AI Cloud}}, \textcolor{blue}{\href{https://www.intel.com/content/www/us/en/developer/tools/devcloud/overview.html}{Intel Developer Cloud}}, and \textcolor{blue}{\href{https://platform.openai.com/tokenizer}{the NegevHPC project}}. Part of this work was completed when Niranjan Hasabnis and Nesreen Ahmed were at Intel Labs and Gal Oren with NRCN.
\end{acks}

\bibliographystyle{ACM-Reference-Format}
\bibliography{references}

\end{document}